\documentclass{article}
\usepackage{spconf,amsmath,epsfig}
\usepackage{multirow}
\usepackage{multicol}
\usepackage{diagbox}
\usepackage{pifont}
\usepackage{subfig}
\usepackage{float}
\usepackage{todonotes}
\restylefloat{figure}
 \definecolor{red}{RGB}{255,0,0}
 \definecolor{green}{RGB}{0,255,0}
 \definecolor{blue}{RGB}{0,0,255}
 \definecolor{cyan}{RGB}{0,255,255}
 \definecolor{magenta}{RGB}{255,0,255}
 \definecolor{lila}{RGB}{153,50,204}
\DeclareGraphicsExtensions{.jpg,.png}
\usepackage{tikz}
\usepackage{booktabs}

\usepackage{enumitem}
\setlist[itemize]{leftmargin=*}
\graphicspath{{./figures}}
 \let\OLDthebibliography\thebibliography
 \renewcommand\thebibliography[1]{
   \OLDthebibliography{#1}
   \setlength{\parskip}{0pt}
   \setlength{\itemsep}{0pt plus 0.3ex}
 }

\title{There Is No Data Like More Data --\\ Current Status of Machine Learning Datasets in Remote Sensing}
\name{Michael Schmitt\textsuperscript{1},  Seyed Ali Ahmadi\textsuperscript{2},  Ronny H{\"a}nsch\textsuperscript{3}}
\address{\textsuperscript{1}Department of Geoinformatics, Munich University of Applied Sciences, Munich, Germany \\
\textsuperscript{2}Faculty of Geodesy and Geomatics Engineering, K. N. Toosi University of Technology, Tehran, Iran\\
\textsuperscript{3}Microwaves and Radar Institute, German Aerospace Center (DLR), Oberpfaffenhofen, Germany
}

\begin{document}
\maketitle
\begin{abstract}
Annotated datasets have become one of the most crucial preconditions for the development and evaluation of machine learning-based methods designed for the automated interpretation of remote sensing data. In this paper, we review the historic development of such datasets, discuss their features based on a few selected examples, and address open issues for future developments.
 
\end{abstract}
\begin{keywords}
Deep Learning, Machine Learning, Datasets, Remote Sensing
\end{keywords}
\section{Introduction}
\label{sec:intro}
In the era of machine learning -- more specifically: deep learning -- the availability of annotated datasets has become one of the most crucial preconditions for the development and evaluation of new methods for the automated interpretation of remote sensing data. While it was possible to train \textit{shallow} learning approaches on comparably small datasets, \textit{deep} learning requires large-scale data to reach the desired generalization performance. 
The main goal of general computer vision is the analysis of every-day images containing every-day objects, such as furniture, animals, or road signs. Thus, extremely large image databases, such as ImageNet\footnote{As a prime example for an annotated computer vision dataset, ImageNet contains more than 14 million images depicting objects from more than 20,000 categories.}, have been created already more than 10 years ago and form the backbone of many modern machine learning developments. In contrast to that, the annotation of remote sensing data is much more complicated due to the dependence on several factors such as sensor technology and target application. To provide a hypothetical example: A dataset for the detection of water surfaces from synthetic aperture radar (SAR) imagery will contain observations and annotations that are very different from the observations and annotations contained in a dataset for the semantic segmentation of urban land cover types from multi-spectral optical data. This lack of generality has led to the generation of uncountable remote sensing datasets. With this paper, we intend to review those developments in order to provide readers with an overview of what is available so far, and what will be needed in the future.

\section{History of Remote Sensing Datasets}
\label{sec:History}
Of course, datasets have always existed in remote sensing. Even before the machine learning era, it was necessary to validate novel signal and image processing algorithms on dedicated test data. The generation -- and publication -- of data dedicated to the training of machine learning algorithms, however has probably only started about 15 years ago, when the IEEE-GRSS Data Fusion Contest was created to foster research in remote sensing data fusion. Already from the second contest (organized in 2007) on, typical machine learning tasks (here: pixel-wise classification of land cover) were in the center of interest. 

\begin{figure*}
    \centering
    \includegraphics[width=\linewidth]{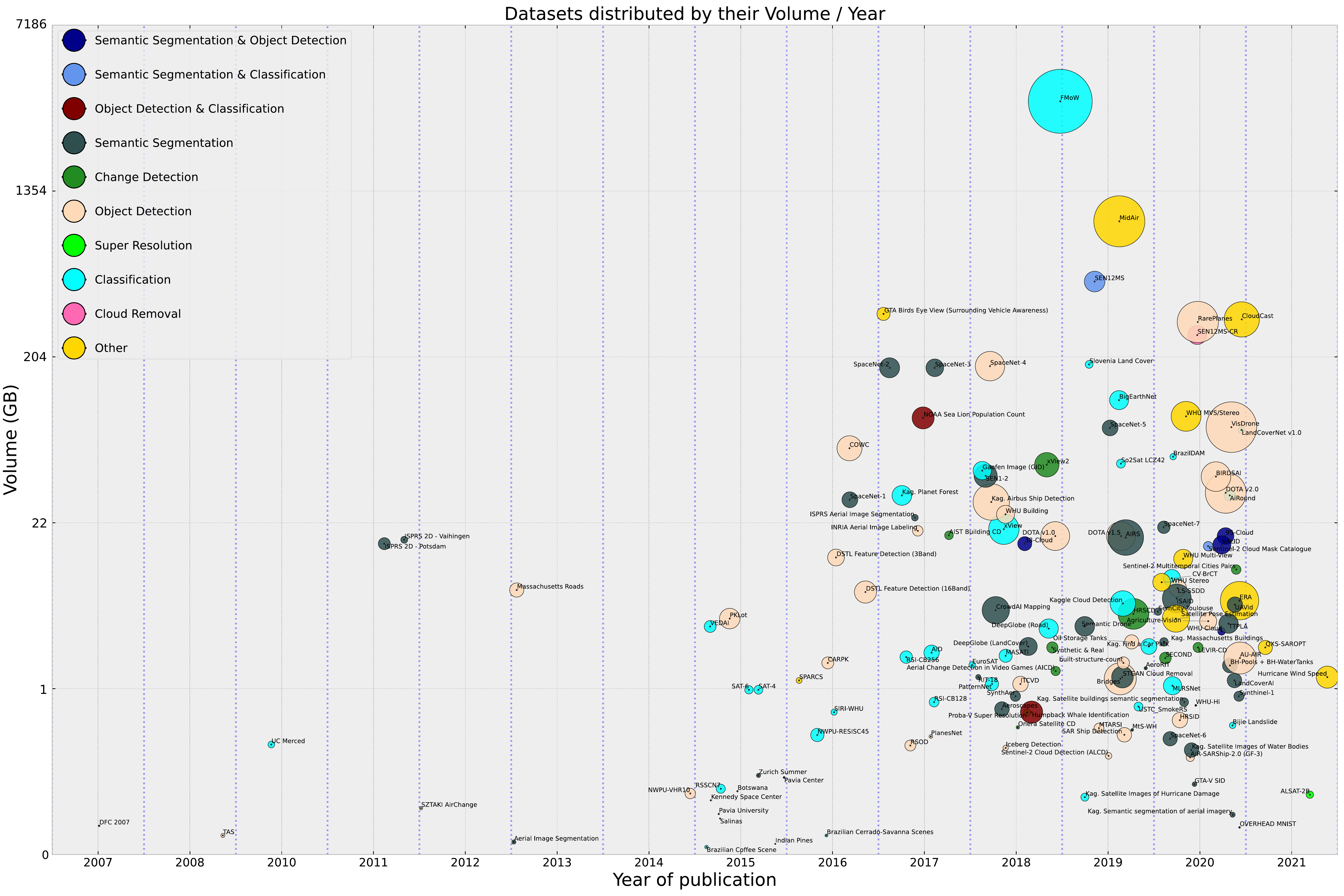}
    \caption{Evolution of remote sensing datasets dedicated to machine learning tasks. Since dataset ``size'' being a hard-to-define measure, it is represented in two ways: The vertical axis relates to the actual data volume, while the circle size relates to the number of spatial pixels covered by the dataset. This way, size is connected to both the spatial dimension as well as the overall information content in terms of implicit features such as resolution, sensors modalities, numbers of bands/channels etc.}
    \label{fig:timeline}
\end{figure*}

The beginnings of ML applied to remote sensing were centered on the analysis of individual study sites. The datasets were comparably small and samples for training, validation, and testing were often taken from the same image. Figure~\ref{fig:timeline} shows the development of datasets over time, illustrating also approximate dataset sizes and purposes. In the context of this paper, we look at the \emph{size} of datasets from two perspectives: 
\begin{enumerate}
    \item \emph{Size} in terms of \emph{spatial pixels}: For this size measure, we count the number of pixels in the highest available image resolution, while ignoring multi-band, multi-channel and multi-sensor data. That is, pixels are only counted once to get a feeling for some form of spatial coverage provided by the dataset.
    \item \emph{Volume} in terms of \emph{data storage}: The amount of disk space required for a dataset serves as an indicator for the provided multitude of modalities (e.g. in the form of multi- or hyperspectral bands or several sensor types), as well as the resolution of the imagery. 
\end{enumerate}

\begin{figure}
    \centering
    \begin{tabular}{c}
     \includegraphics[width=0.8\linewidth]{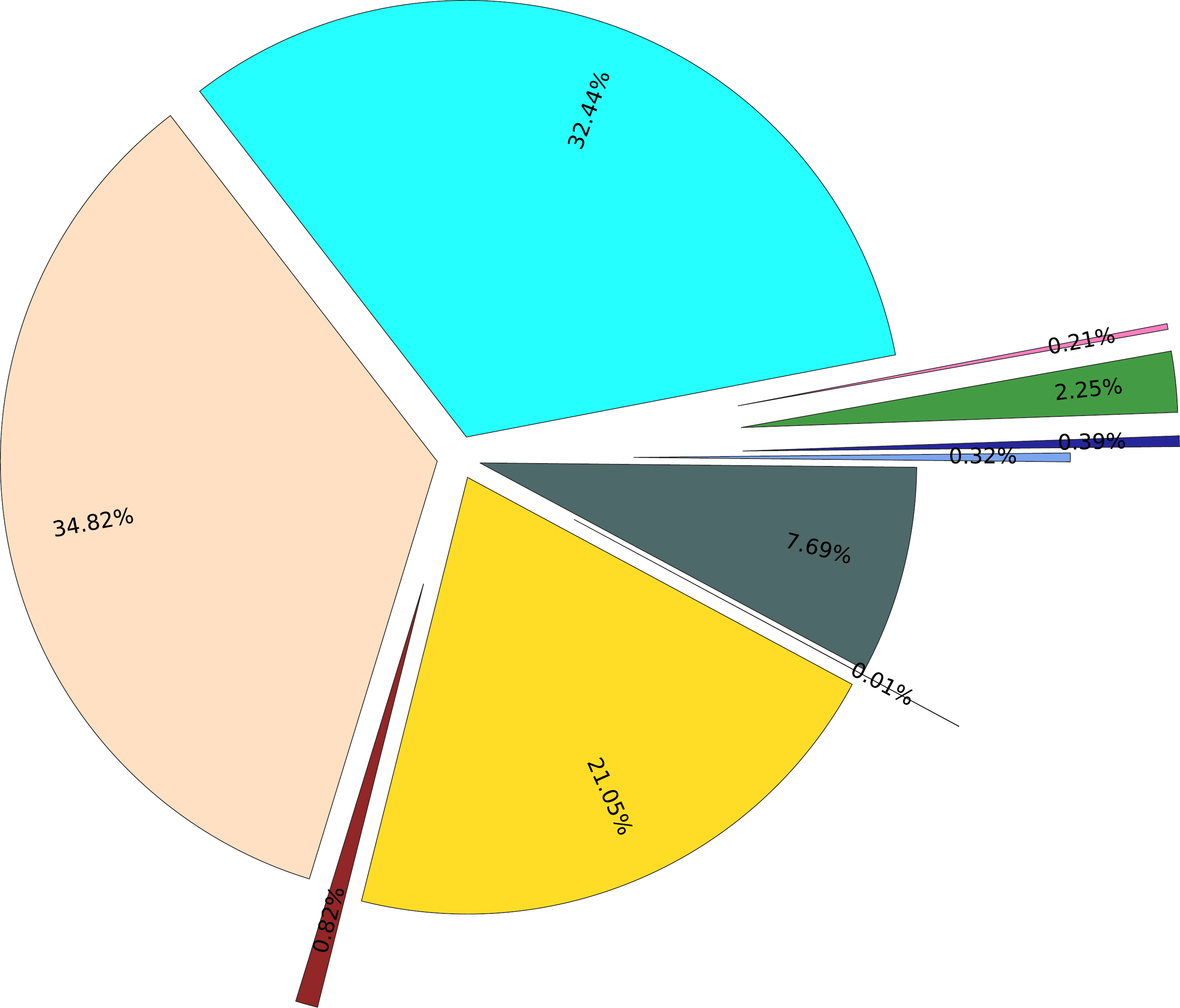}\\
    (a)\\
    \includegraphics[width=0.8\linewidth]{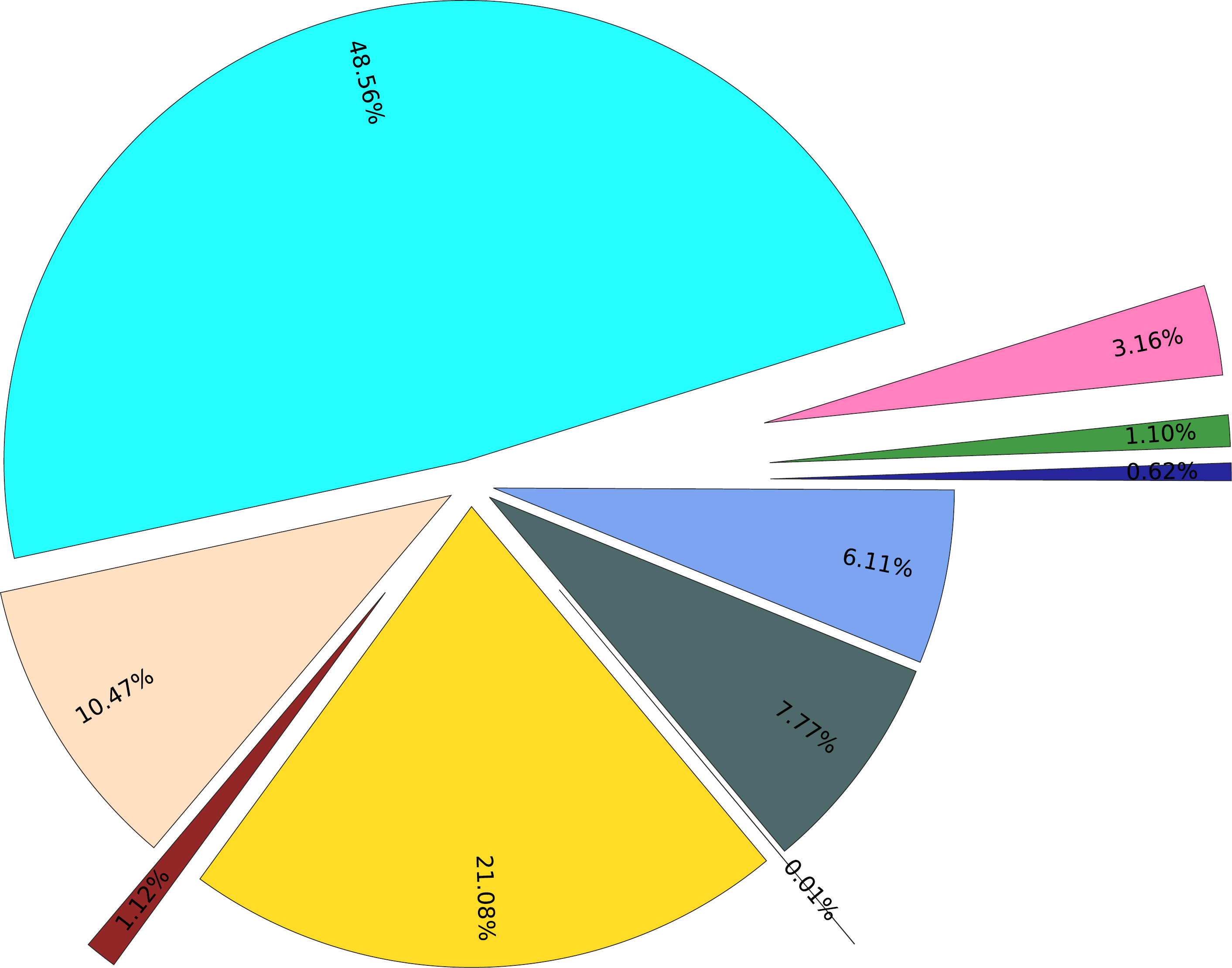}\\
    (b)\\
    \includegraphics[width=0.6\linewidth]{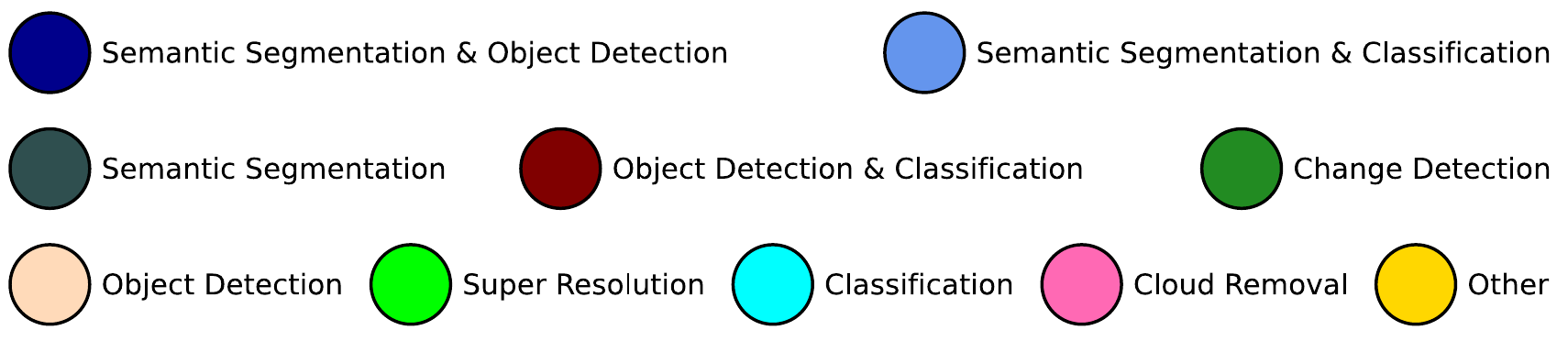}
    \end{tabular}
    \caption{Distribution of dataset sizes over the typical remote sensing tasks: (a) Dataset size expressed in number of pixels; (b) dataset size expressed in data volume.}
    \label{fig:distribution}
\end{figure}

\noindent In this context, it is important to mention that we have tried our best to collect information about as many datasets as possible. However, it is clear that there will always be datasets we have not yet become aware of. Besides, for some datasets, we were not able to gather all required information. In spite of that, we believe that several interesting insights can be drawn from the timeline:
\begin{itemize}
    \item Besides the data provided in the frame of the IEEE GRSS Data Fusion Contests, there are a few other pioneering datasets, which have certainly fostered research of machine learning applied to remote sensing data in its early stages. Those are the UC Merced dataset as the first dataset dedicated to scene classification \cite{Yang2010}; the ISPRS Vaihingen/Potsdam dataset, which was originally intended to benchmark semantic segmentation approaches tailored to aerial imagery \cite{Rottensteiner2012}, but has also been used to train methods for other tasks, e.g. single-image height reconstruction in the meantime; and the SZTAKI-INRIA dataset designed for object detection \cite{Benedek2011}.
    \item While more and more datasets have been released starting from 2015, 2018 seems to be the year from which on datasets grew larger and larger both in terms of spatial extent and multi-modal information content. 
    \item As also confirmed by the statistics displayed in Fig.~\ref{fig:distribution}, most pixels are available for the task of object detection, which correlates to its popularity in deep learning-oriented remote sensing research. When it comes to data volume, however, semantic segmentation and classification lead the way, which indicates that in contrast to object detection here more multi-modal data are used. 
    \item As of the time of writing of this paper, we were aware of 181 datasets aiming at the combination of machine learning and remote sensing, of which we were able to consider 141 due to all necessary information being available. All of those datasets provide different features, address different sensors, different resolutions and different tasks. Thus, there is not yet \emph{the} single go-to dataset that is used for pre-training most newly developed models, or for benchmarking certain tasks against the state-of-the-art. 
\end{itemize}

\section{Two Important Examples}
In this section, we describe both the oldest/first and the largest currently available remote-sensing oriented machine learning datasets  to provide examples giving a more detailed view of the peculiarities of such datasets and the developments during the last years.

\subsection{DFC2007}
As mentioned above, to our knowledge the data provided in the frame of the 2007 IEEE-GRSS Data Fusion Contest was the first dataset at the intersection of remote sensing and machine learning that was openly published for the benefit of the community \cite{Pacifici2008}. It is a typical old-school dataset sampled from just a single study scene (namely the city of Pavia, Italy). It contains two multi-spectral Landsat images (acquired in 1994 and 2000), as well as a time series of 9 ERS SAR images (acquired between August 1992 and July 1995), as well as a sparse map annotating several pixel patches into four generic urban land cover classes. Due to the limitation to one scene, the dataset enables the investigation of scene-specific models, which can either be from classical image processing or employ shallow learning. 

\subsection{Functional Maps of the World (FMoW)}
According to our timeline in Fig.~\ref{fig:timeline}, the Functional Maps of the World dataset \cite{Christie2018} currently is the largest available dataset both in terms of pixels and data volume. It consists of 1{,}047{,}691 images from 207 countries and is made for the development of machine learning models for the prediction of the functional purposes of buildings and land use from temporal sequences of satellite images and corresponding metadata features  about location, time, sun angles, physical sizes etc. All the image data of FMoW stem from the Digital Globe  constellation and were gathered in pairs, consisting of 4-band (Quickbird-2 or GeoEye-1) or 8-band (Worldview-2/3) multispectral imagery in the visible to near-infrared region, as well as a pan-sharpened RGB image that represents a fusion of the high-resolution panchromatic  image  and  the  RGB  bands  from  the  lower-resolution  multispectral  image.

\section{Current Status and Open Issues}
As mentioned in Section~\ref{sec:History}, the amount of datasets for machine learning in remote sensing continues to grow, as new tasks and new sensors, combined with ever-improving possibilities to handle big geospatial data, require new materials for training and evaluating new solutions. On the downside, a one-for-all go-to solution -- a remote sensing-oriented \emph{ImageNet} -- is still not in sight. While this is not really a problem with respect to the generation of sensor- and task-specific models, it introduces a significant overhead in terms of data preparation for every new undertaking in algorithm development; and it also hinders the comparability of newly developed methods, as transparent benchmarks do not really exist. We thus hope that the future will bring joint endeavors aiming at the establishment of such a standard database. This could either be realized from scratch, or build upon one of the larger existing datasets. In any case, it is important that this standard database meets the following criteria:
\begin{itemize}
    \item Its data should be sampled across the globe and throughout the year to cover as many cultural and climatic regions as possible.
    \item The dataset should contain data from as many modalities as possible. Starting with freely available satellite imagery, this refers to at least the different Sentinel satellites and the Landsat mission. Of course, a possibility to add higher resolution satellite or even aerial data would be highly desirable.
    \item Ideally, the dataset would try to cover several remote sensing tasks. Instead of just focusing on object detection or scene classification, a multi-use annotation would enhance re-usability significantly.
    \item Since manual labeling of large amounts of remote sensing imagery is very expensive, time-consuming, difficult, and error-prone, robust globally transferable land cover schemes that address multiple semantic scales have to be defined. In addition, research has to be invested in reliable ways to source the required annotations. Options include automated labeling from existing geodata as well as crowd-sourced mapping -- or a combination thereof.
\end{itemize}

\section{Summary and Conclusion}
In this paper, we have summarized the developments in datasets for machine learning applied to remote sensing problems. We have shown the historic timeline, addressed the pecularities of remote sensing data illustrated by some example datasets, and discussed the current status and open issues. While an increase in the availability of annotated data is observable, due to the heterogeneity of remote sensing measurements and tasks, there is still not a single go-to dataset, which could serve the purpose of transparent benchmarking and standardized pre-training. However, the increasing importance of open data satellite missions such as the Sentinels of the European Copernicus program or ``AI for social good''-based open data initiatives by companies such as Microsoft pave the way to fill this gap in the future. Ideally, existing large-scale datasets are built upon and extended for this purpose to benefit from existing expertise.

\section*{\normalfont\normalsize\bfseries\centering{\MakeUppercase{Acknowledgements}}}
Michael Schmitt would like to acknowledge support by the Open Data Impact Award 2020 from the German Stifterverband.

% -------------------------------------------------------------------------
\bibliographystyle{IEEEbib}
\bibliography{References}

\end{document}